\documentclass[11pt]{article}

\usepackage[final]{acl}

\usepackage{times}
\usepackage{latexsym}

\usepackage[T1]{fontenc}

\usepackage[utf8]{inputenc}

\usepackage{microtype}
\usepackage{amsmath}
\usepackage{inconsolata}

\usepackage{graphicx}
\usepackage{multirow}
\usepackage{amssymb}
\usepackage[dvipsnames]{xcolor}

\definecolor{cvprblue}{rgb}{0.21,0.49,0.74}

\usepackage{marvosym}
\usepackage{makecell}
\usepackage{xcolor}
\usepackage[table]{xcolor}
\usepackage{booktabs}
\usepackage{tabularx}
\definecolor{nblue}{cmyk}{0.95,0.0,0.2,0.2}
\definecolor{green}{HTML}{009B55}
\definecolor{githubcolor}{HTML}{181717}
\definecolor{huggingfacecolor}{HTML}{FFD21E}

\usepackage{simpleicons}
\ifPDFTeX
  \pdfmapfile{+simpleicons.map}
\fi

\usepackage{pifont}
\providecolor{cmarkgreen}{HTML}{2E9E5B}
\providecolor{xmarkred}{HTML}{D64545}
\providecolor{pmarkamber}{HTML}{E39B2D}
\def\githuburl{https://github.com/IQuestLab/RepoFactory}
\def\huggingfaceurl{https://huggingface.co/datasets/IQuestLab/RepoFactory}
\newcommand{\resourceLink}[2]{%
  \ifx#1\empty
    #2%
  \else
    \href{#1}{#2}%
  \fi
}

\title{
 \vspace{-45pt}
  \makebox[\linewidth][l]{\includegraphics[height=0.75cm]{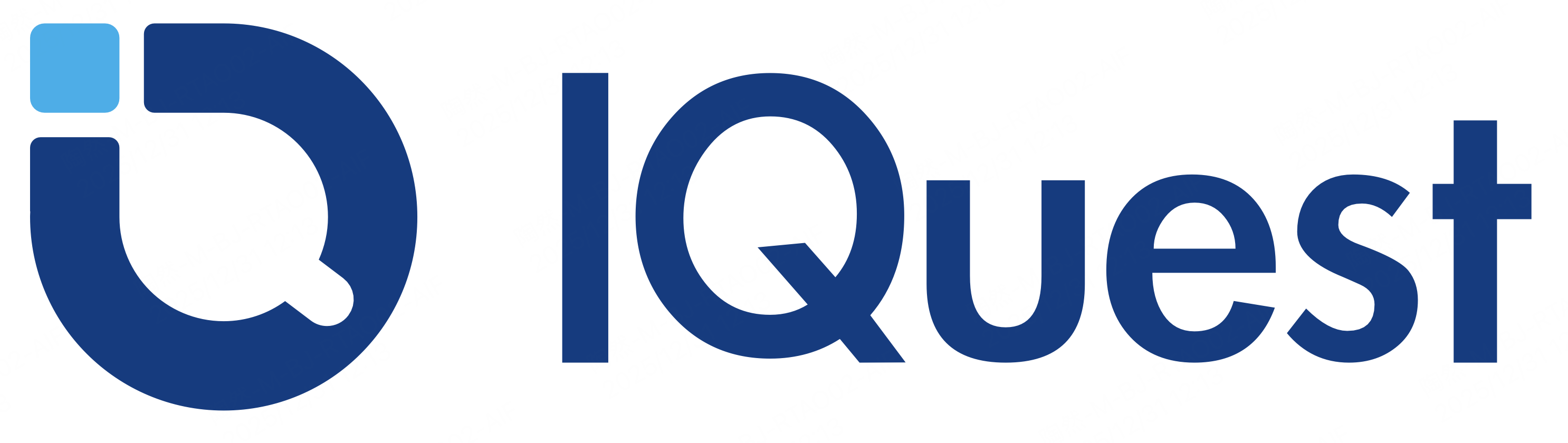}}\\[-10pt]
  \rule{\linewidth}{0.1pt}\\[5pt]
From Repositories to Training Environments: \\Scaling Long-Horizon Coding Agents with Environment Evolution Loop
}

\author{
\begin{tabular}{@{}c@{}}
\textbf{Yang Tian}$^{2,3}$ \quad\,\,\,\,
\textbf{Tianyu Zheng}$^{3,*}$ \quad\,\,\,\,
\textbf{Xiyuan Yang}$^{2,3}$ \quad\,\,\,\,
\textbf{Yaxin Du}$^{2,3}$ \\
\textbf{Chengran Yang}$^{3}$ \quad\,\,\,\,
\textbf{Bo Zhao}$^{2}$ \quad\,\,\,\,
\textbf{Jian Yang}$^{1,\dagger}$ \quad\,\,\,\,
\textbf{Bryan Dai}$^{3}$
\end{tabular}
\\[1em]
\begin{tabular}{@{}c@{}}
$^{1}$Beihang University \quad
$^{2}$School of AI, Shanghai Jiao Tong University \quad
$^{3}$IQuest Research
\end{tabular}
\\[0.5em]
\textit{$^*$ Project Leader \qquad
$^\dagger$ Corresponding Author}
\\[0.5em]
\resourceLink{\githuburl}{\textcolor{githubcolor}{\simpleicon{github}}\,GitHub}
\quad
\resourceLink{\huggingfaceurl}{\raisebox{-0.15\height}{\includegraphics[height=1.15em]{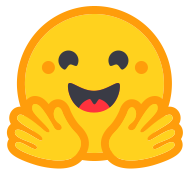}}\,Hugging Face}
}

\begin{document}
\maketitle

\begin{abstract}
Training coding agents for long-horizon tasks requires environments with verifiable feedback. Existing pipelines often follow a linear process that ends with validation, establishing usability but providing little guidance on improving training value. Recursive self-improvement (RSI) often requires model rollouts or model training followed by downstream evaluation, with costs and feedback delays amplified on long-horizon tasks. How can we use timely, reliable proxy signals to improve training value and synthesize environments at scale? We present \textbf{RepoFactory}, an automated pipeline that converts real repositories into verifiable environments where agents build complete projects from scratch given a language specification. Its Environment Evolution Loop optimizes a multiplicative objective combining validity, structural coupling, and implementation scope through proxy feedback. RepoFactory produces 10,159 environments across 17 software domains. Training Qwen3.5-35B-A3B on these environments yields RepoHorizen-35B-A3B, which achieves gains of 22.6 percentage points on NL2Repo-Bench and 8.5 points on average across five benchmarks covering long-horizon coding and agentic tasks in real-world production settings, compared with 17.5 and 6.4 points without environment evolution. These results highlight the training value of RepoFactory environments and the additional gains from the Environment Evolution Loop.
\end{abstract}

\section{Introduction}
\label{sec:intro}

Large language models (LLMs) are increasingly being developed into agents that can carry out consequential work in real-world environments~\cite{glm2026glm,xu2026autolab,team2026ai4ai,yang2026iquest,deepseek2606deepseek}. Unlike static language tasks, agentic work unfolds through sustained interaction with environments. This transition changes the objective of model training: a capable agent must learn not only to produce correct answers, but also to make coherent progress toward completing long-horizon tasks~\cite{li2026recursive,tian2026beyond,bajoria2026language}.

Software engineering has become a major application of coding agents. Pipelines built around bug fixing and feature development have provided verifiable tasks for training agents on real codebases~\cite{SWE-Gym,SWE-smith}. A successful local change, however, is only part of the work involved in delivering software. Coding agents are increasingly expected to handle more complex development processes and produce complete, working systems~\cite{icoder,team2026kimi}. Training on short-horizon tasks alone cannot cover this broader process, making it increasingly important to build repository-level environments for agent training.

Constructing environments for long-horizon tasks at the repository level remains challenging. \textbf{(i) Rigid construction pipelines limit the usable repository pool.} DeNovoSWE~\cite{zhao2026denovoswe} and MindForge~\cite{chen2026mindforge} follow linear construction pipelines. They refine documentation or resolve build failures while keeping the reference implementation fixed. Repositories that remain unsuitable are discarded, even when further revision could make them usable. Scaling therefore depends on finding more repositories that meet the same fixed requirements. \textbf{(ii) Feedback focused on reliability overlooks training value.} RefactorForge~\cite{ebrahimi2026software} and TrajLong~\cite{peng2026trajlong} design data for specific capabilities but do not explore how environment feedback can improve task content. ProgramGym~\cite{yang2026programgym} uses execution diagnostics and coverage feedback to repair builds and strengthen tests around a fixed reference implementation. These signals improve reliability without explicitly optimizing training value, which may remain limited when a task requires little implementation work or coordination across components. \textbf{(iii) Feedback from model execution is costly and time-consuming.} BenchEvolver~\cite{BenchEvolver} and SETA~\cite{seta} use model performance to guide task revision. Obtaining this feedback requires model rollouts on the candidate tasks. For long-horizon coding tasks, each rollout can involve extensive implementation, increasing computational costs and delaying subsequent revisions. How to obtain timely, reliable feedback at low cost to improve training value and synthesize long-horizon coding environments at scale remains an open research question.

We present RepoFactory, an automated pipeline that transforms real-world repositories into verifiable environments. We design these environments around full repository generation task, in which an agent must build a complete, functioning repository from scratch given a language specification. We define a multiplicative objective combining validity, structural coupling, and implementation scope. It provides timely feedback at low cost without requiring a model to solve each candidate task. Our Environment Evolution Loop uses this objective to guide joint revisions to the oracle implementation, task specification, and tests. Validity feedback helps the loop repair initially unsuitable environments and recover repositories that would otherwise be discarded. Structural coupling and implementation scope serve as proxies for training value, directing revisions toward the implementation and integration work required by the task. A construction agent chooses how to modify each candidate and receives updated measurements after each revision, iterating within a fixed construction budget. Our main contributions are threefold:


\begin{itemize}
    \item We introduce RepoFactory, an automated pipeline that converts real-world projects into 10,159 verifiable full-repository generation environments across 17 software domains.
    \item We develop an Environment Evolution Loop that optimizes a multiplicative objective combining validity, structural coupling, and implementation scope. Timely, low-cost proxy feedback guides improvements to both environment reliability and training value.
    \item Training Qwen3.5-35B-A3B on RepoFactory yields RepoHorizen-35B-A3B, with gains of 22.6 percentage points on NL2Repo-Bench and 8.5 points on average across five coding and agentic benchmarks. Environment evolution contributes 5.1 and 2.1 points to these gains, respectively.
\end{itemize}

\section{Related Work}
\label{sec:related_work}

\subsection{Scaling Verifiable Environments}

Environment synthesis scales agent training by generating interactive tasks with executable verification~\cite{Autodata,Framework,Training}. EnvScaler~\cite{EnvScaler} and ScaleEnv~\cite{ScaleEnv} target tool use, while Endless Terminals~\cite{Endless}, LiteCoder-Terminal~\cite{LiteCoder-Terminal}, and CLI-Universe~\cite{CLI-Universe} extend this approach to terminal-agent training. Broader task coverage, however, does not ensure that tasks preserve the interdependent requirements of real software or demand sustained development. RepoFactory derives full-repository generation tasks from real projects, grounding task behavior in existing implementations and tests. Reconstructing an entire repository requires sustained implementation and integration across these interdependent components.



\subsection{Recursive Synthesis and Agent Self-Improvement}

Self-training reuses successful solutions as supervision, while self-play generates new problems for agents to solve~\cite{Self-Play,MM-Zero}. BenchEvolver~\cite{BenchEvolver} evolves reference solutions to construct harder tasks, while SETA~\cite{seta} adapts terminal environments using learner performance. DataMaster~\cite{DataMaster} optimizes training data using downstream evaluation scores, adding model training to the feedback cycle. These approaches rely on solver rollouts or downstream training, increasing computational costs and delaying feedback. To our knowledge, RepoFactory is the first to evolve full-repository generation environments using an objective that combines validity checks with inexpensive code metrics. This objective provides timely feedback without requiring solver rollouts or downstream training.

\begin{figure*}[t]
  \centerline{\includegraphics[width=0.98\textwidth]{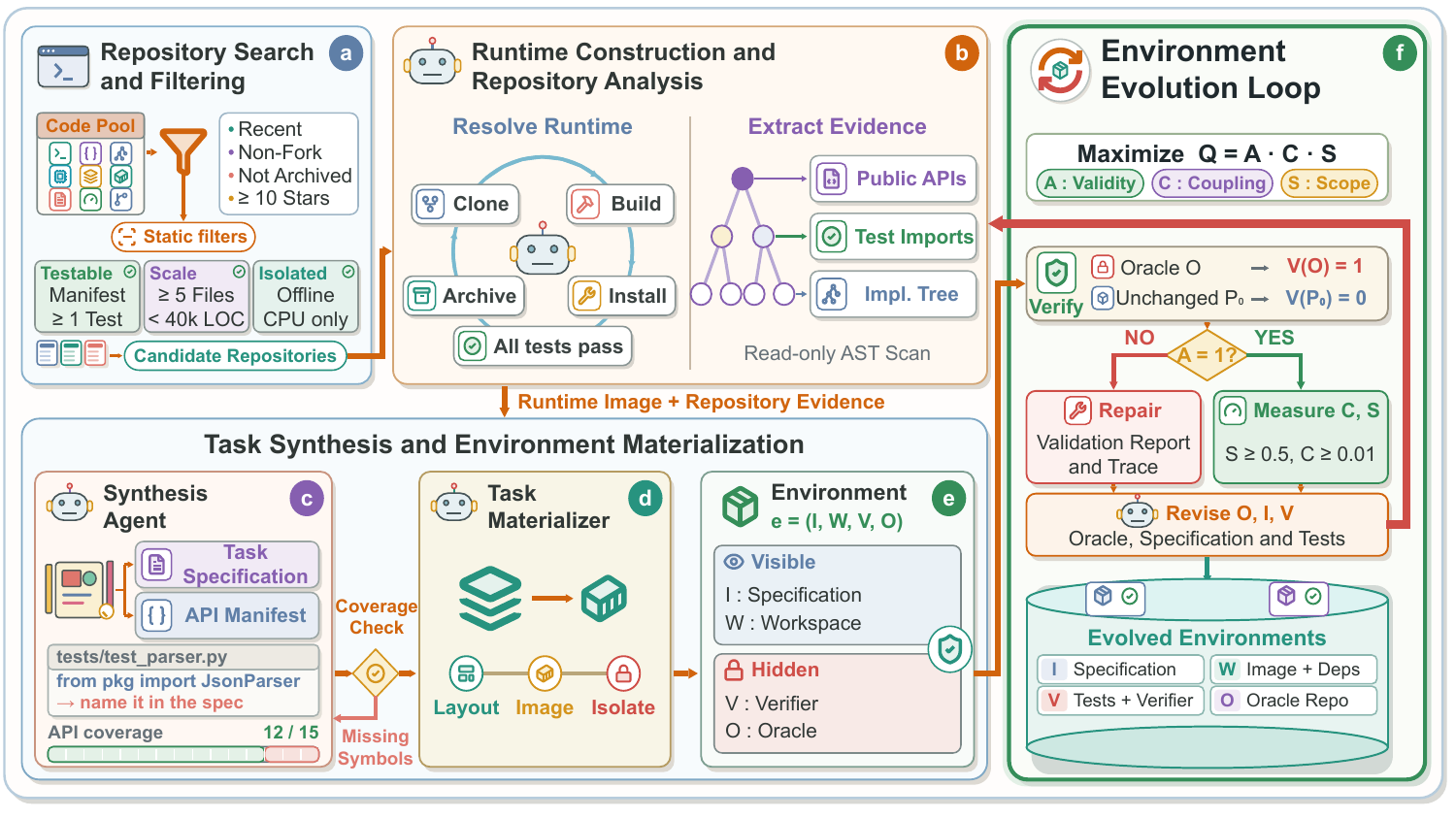}}
\caption{Overview of RepoFactory. The pipeline searches for repositories, reconstructs their runtimes, and synthesizes task specifications from repository evidence. The Environment Evolution Loop uses low-cost feedback to improve environment reliability and training value.}
  \label{main}
\end{figure*}

\section{RepoFactory}
\label{sec:method}

RepoFactory converts real-world repositories into verifiable environments where agents reconstruct complete projects from a language specifications. The Environment Evolution Loop uses timely, low-cost feedback from verification and code analysis to jointly revise oracle implementations, task specifications, and tests, optimizing a multiplicative objective over validity, structural coupling, and implementation scope. All construction agents use Qwen3.8-Max. Figure~\ref{main} shows the overall pipeline.


\subsection{Problem Formulation and Objective}
\label{sec:problem-objective}

Our goal is to transform a candidate repository $r$ into a training environment $e=(I,W,V,O)$, where $I$ is the language specification, $W$ is an isolated workspace and runtime, $V$ is the verifier, and $O$ is the source repository's oracle implementation. During training, an agent $\pi$ receives $(I,W)$ and reconstructs a project $P=\pi(I,W)$, which is scored by $V(P)\in[0,1]$.

We define the environment objective as
\begin{equation}
    Q=A\cdot C\cdot S,
    \label{eq:environment-objective}
\end{equation}
where $A$, $C$, and $S$ measure validity, structural coupling, and implementation scope, respectively. The validity term requires full reward for the oracle and zero reward for the unchanged workspace $P_0$:
\begin{equation}
    A=
    \begin{cases}
        1, & V(O)=1\ \text{and}\ V(P_0)=0,\\
        0, & \text{otherwise}.
    \end{cases}
    \label{eq:environment-validity}
\end{equation}

We measure structural coupling with normalized propagation cost:
\begin{equation}
    C=\frac{R}{N^2},
    \label{eq:structural-coupling}
\end{equation}
where $N$ is the number of functions in the analyzed oracle code and $R$ is the number of reachable ordered function pairs in its static internal call graph, including self-pairs.

Implementation scope combines code size and dependency breadth:
\begin{equation}
    S = \log_{10}(\mathrm{LOC}) + \log_{10}(1 + D) - 2,
    \label{eq:implementation-scope}
\end{equation}
where $\mathrm{LOC}$ is the source code size and $D$ is the number of direct runtime dependencies.

All three terms are evaluated on the current candidate. The Environment Evolution Loop iteratively revises the environment to maximize $Q$ within a fixed construction budget.



\subsection{Construction Pipeline}
\paragraph{Repository Search and Filtering}
RepoFactory searches public code-hosting platforms for candidate repositories, excluding those included in established coding benchmarks to avoid evaluation overlap. Eligible projects must provide package metadata and a nonempty local test suite to support isolated installation and evaluation. To balance task complexity with reconstruction cost, we require at least five implementation files and fewer than 40,000 lines of Python code. We exclude projects that require external services, outbound network access, graphical interfaces, or specialized hardware. Network-related projects remain eligible when their tests mock external calls or use only local connections.

\paragraph{Runtime Construction and Repository Analysis}
For each candidate repository, RepoFactory copies each candidate repository into an isolated task directory and builds a Docker image from its build files and dependency manifests. A agent installs dependencies, identifies test commands, and revises the image or execution configuration as needed. We retain a repository only when all tests pass in a fresh container with bounded resources and no network access. RepoFactory archives the validated image and records the installation and test commands, package files, and test results.

After establishing the runtime, RepoFactory analyzes the repository in a read only container to determine what the task specification must cover. A scanner based on abstract syntax trees extracts public classes, functions, and methods, together with their signatures, exceptions, and return information. It also records repository symbols imported by tests and builds an implementation tree excluding test files. These outputs guide task synthesis and, together with dependency metadata, provide the inputs for computing structural coupling and implementation scope.


\paragraph{Task Synthesis and Environment Materialization.}
RepoFactory prepares an isolated workspace with dependencies installed and disabled external network access. The coding agent must reconstruct the project from the task specification, without access to the original repository or evaluation code. The synthesis agent uses the repository analysis to write this specification and an accompanying API manifest. The specification describes the repository's functionality and structure, while the API manifest documents the public interfaces the agent must implement.

Functionally correct implementations can still fail if their interface names or import paths differ from those expected by the tests. RepoFactory therefore checks the specification and API manifest against the repository symbols imported by the original tests, returning missing symbols and their import records to the synthesis agent for revision. Once this check passes, RepoFactory combines the specification with the validated runtime image to create the environment. The original repository remains isolated as the oracle, and evaluation resources are available only during verification.

\paragraph{Environment Evolution Loop}
After materialization, RepoFactory first checks validity $A$. The oracle $O$ must receive full reward, and the unchanged initial workspace must receive zero reward. If either condition fails, the construction agent uses the validation report and execution trace to identify the cause and repair the candidate. The revised environment is evaluated again, and optimization of scope and coupling begins only after both checks pass.

A valid environment may still require little implementation work. Implementation scope $S$ therefore guides revisions toward more substantial reconstruction tasks. A larger project can still consist of independent components, so structural coupling $C$ complements scope by directing evolution toward more interdependent functionality. The construction agent decides how to revise the oracle implementation, task specification, and tests, keeping the description and evaluation consistent with the revised functionality.

For each valid candidate, the agent compares $S$ and $C$ against thresholds of $0.5$ and $0.01$, respectively, and prioritizes any term below threshold. When both thresholds are met, the agent chooses which term to optimize under $Q$. Each revision is rechecked for validity before the scores are reassessed, with validity failures returned to repair. A revision targeting $S$ or $C$ is retained only if the selected score strictly increases; otherwise, RepoFactory restores the previous version. The loop runs for up to 12 rounds, assessing each revision through verifier execution and code analysis without solving-agent rollouts or downstream training and evaluation. RepoFactory uses this low-cost feedback to recover failed candidates and improve valid tasks, increasing both the yield and training value of repository-level environments.

\begin{figure*}[t]
  \centerline{\includegraphics[width=0.95\textwidth]{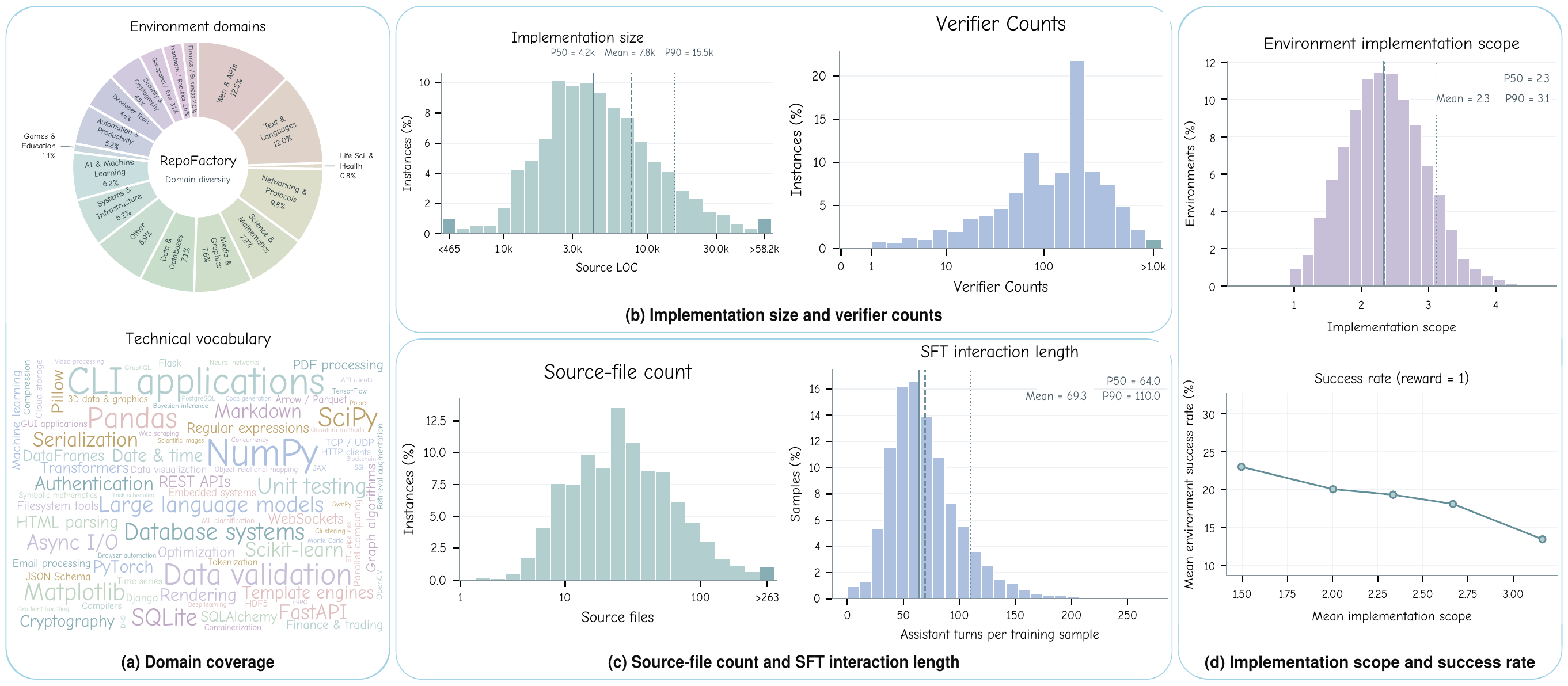}}
\caption{Statistics of RepoFactory environments and SFT data.}
  \label{fig:data_analysis}
\end{figure*}
\section{Analysis of Synthesized Data}
\label{sec:data_analysis}
Effective training environments for long-horizon coding should expose agents to diverse software and require substantial repository reconstruction under reliable feedback across a broad range of difficulty. Our 10,159 RepoFactory environments span 17 domains and place agents in complete repository reconstruction tasks with executable validation. To collect SFT data, we run Qwen3.8-Max five times on each environment, filter out low reward trajectories, and retain 10,947 samples spanning 5,577 environments. We analyze the resulting collection to characterize its diversity, scale, interaction length, and difficulty with the statistics summarized in Figure~\ref{fig:data_analysis}.

\paragraph{Diversity.}
The collection covers a broad range of application domains and technical ecosystems, without being dominated by any single area, as shown in Figure~\ref{fig:data_analysis}(a). The largest category accounts for only 12.5\% of environments, and only 6.9\% fall outside the named domains, indicating broad coverage rather than concentration. Its varied libraries, interfaces, and implementation conventions expose agents to the diversity of real software development. This breadth supports training agents that generalize across software projects instead of specializing in a narrow domain.

\paragraph{Scale and verification.}
These environments require agents to reconstruct substantial codebases rather than write isolated programs. As shown in Figure~\ref{fig:data_analysis}(b), the median oracle contains 4.2k lines of source code, with a substantial tail of larger systems. Across the 10,159 instances with successful test parsing, the verifier contains 2.09 million statically identified test definitions, providing dense coverage at the instance level. Because these tests are inherited from the source repositories, they exercise behavior across the public interface rather than checking only a few end to end outcomes. Success therefore depends on implementing a coherent repository whose components remain consistent under extensive verification.


\paragraph{Repository Scale and Interaction diversity.}
The environments and trajectories capture sustained development at the repository level. The median oracle repository contains 26 source files, and the median SFT trajectory contains 64 assistant turns, as shown in Figure~\ref{fig:data_analysis}(c). A successful trajectory must keep track of a multi-file project while successive edits are implemented and tested. Together, the long-horizon trajectories and varied operations provide rich supervision for maintaining project state as agents coordinate changes across files and refine their implementations through execution feedback.

\paragraph{Implementation Scope.}
Figure~\ref{fig:data_analysis}(d) shows the distribution of implementation scope $S$ and its relationship with reconstruction performance. We measure each environment's success rate as the fraction of teacher trajectories receiving full reward. Environments with matched trajectories are ordered by $S$ and divided into five equally sized groups, whose mean scope and success rates are plotted in the lower panel. Mean success falls from 23.0\% in the group with the smallest scope to 13.4\% in the group with the largest. This trend supports $S$ as a proxy for the demands of repository reconstruction.


\begin{table*}[htbp]
    \centering
    \begingroup
    \setlength{\tabcolsep}{2pt}
    \renewcommand{\arraystretch}{1.0}
    \fontsize{7.5}{8.5}\selectfont
    \resizebox{0.96\linewidth}{!}{%
    \begin{tabular}{l*{7}{c}}
        \toprule
        \textbf{Model}
        & \textbf{Params}
        & \shortstack{NL2Repo\\Bench}
        & \shortstack{SWE-bench\\Verified}
        & \shortstack{SWE-bench\\Pro}
        & \shortstack{SWE-bench\\Multilingual}
        & \shortstack{Terminal-\\Bench 2.1}
        & \textbf{Avg.} \\
        \midrule

        \rowcolor{black!5}
        \multicolumn{8}{l}{\strut\textbf{Frontier Closed-Source Models}} \\

        GPT-5.6-sol & - & 56.8 & 96.2 & 64.6 & - & 88.8 & 76.6 \\
        GPT-5.5 & - & 45.1 & 82.6 & 58.6 & 77.8 & 73.8 & 67.6 \\
        Claude Opus 5 & - & 75.3 & 96.0 & 79.2 & 89.5 & 89.1 & 85.8 \\
        Gemini-3.1 Pro & - & 33.4 & 80.6 & 54.2 & 44.0 & 70.3 & 56.5 \\

        \midrule

        \rowcolor{black!5}
        \multicolumn{8}{l}{\strut\textbf{Open-Weight Models ($>$200B)}} \\

        Qwen3.8-Max & 2.4T-A95B & 55.9 & 85.6 & 67.7 & - & 86.6 & 74.0 \\
        DeepSeek-V4-Pro & 1.6T-A49B & 38.5 & 80.6 & 55.4 & 76.2 & 78.7 & 65.9 \\
        GLM-5.2 & 744B-A40B & 48.9 & 82.8 & 62.1 & 74.5 & 81.0 & 69.9 \\
        Kimi-K2.6 & 1T-A32B & - & 80.2 & 58.6 & 76.7 & 67.2 & 70.7 \\
        Tencent-Hy3 & 295B-A21B & 45.6 & 78.0 & 57.9 & 75.8 & 71.7 & 65.8 \\
        MiniMax-M3 & 428B-A23B & 42.1 & 80.5 & 59.0 & - & 66.0 & 61.9 \\

        \midrule

        \rowcolor{black!5}
        \multicolumn{8}{l}{\strut\textbf{Open-Weight Models ($<$200B)}} \\

        Qwen3.5-122B-A10B & 122B-A10B & 27.7 & 73.4 & 49.5 & 67.2 & 44.9 & 52.5 \\
        Ornith-1.0-35B & 35B-A3B & 34.6 & 75.6 & 50.4 & 69.3 & 62.8 & 58.5 \\

        Qwen3.5-35B-A3B
        & 35B-A3B
        & 23.8
        & 65.8
        & 42.4
        & 59.7
        & 33.7
        & 45.1 \\

        \midrule

        \rowcolor[HTML]{DFE7F5}
        \textbf{Ours} & & & & & & & \\

        RepoHorizen-35B-A3B
        & 35B-A3B
        & 46.4\,\textcolor{green!50!black}{(+22.6)}
        & 72.2\,\textcolor{green!50!black}{(+6.4)}
        & 47.6\,\textcolor{green!50!black}{(+5.2)}
        & 61.0\,\textcolor{green!50!black}{(+1.3)}
        & 40.5\,\textcolor{green!50!black}{(+6.8)}
        & 53.5\,\textcolor{green!50!black}{(+8.5)} \\

        \quad \textit{w/o environment evolution}
        & 35B-A3B
        & 41.3\,\textcolor{green!50!black}{(+17.5)}
        & 70.9\,\textcolor{green!50!black}{(+5.1)}
        & 45.1\,\textcolor{green!50!black}{(+2.7)}
        & 60.8\,\textcolor{green!50!black}{(+1.1)}
        & 39.3\,\textcolor{green!50!black}{(+5.6)}
        & 51.5\,\textcolor{green!50!black}{(+6.4)} \\

        \bottomrule
    \end{tabular}%
    }
    \endgroup
    \caption{Main results on five agent benchmarks.}
    \label{tab:main-results}
\end{table*}

\section{Experiments}

\subsection{Experimental Setup}

\paragraph{Benchmarks.}
We evaluate our model on five benchmarks covering repository generation, software engineering, and terminal-based tasks. NL2Repo~\cite{NL2Repo-Bench} is a long-horizon coding benchmark that requires agents to implement a complete repository from scratch based on a natural language description. Success requires sustained planning, tool use, and coordinated implementation across multiple files and modules. SWE-bench Verified~\cite{SWE-bench}, SWE-bench Pro~\cite{SWE-Bench-pro}, and SWE-bench Multilingual~\cite{Multi-SWE-bench} assess the ability to resolve issues in existing repositories, while Terminal-Bench 2.1~\cite{Terminal-bench} evaluates task completion in terminal environments. Together, these benchmarks measure coding-agent performance across both project construction and repository maintenance.

\paragraph{Evaluation Settings.}
We use Claude Code version 2.1.140 as the agent scaffold. The maximum context length is 262,144 tokens, and the maximum output length is 32,768 tokens. Context is automatically compacted when its usage reaches 80\% of the maximum context length. For decoding, we set temperature to 1.0, top-$p$ to 0.95, and top-$k$ to 20.

\subsection{Main Results}
Table~\ref{tab:main-results} compares RepoHorizen-35B-A3B and its base model with representative closed-source and open-weight models. On NL2Repo, which directly evaluates full-repository reconstruction, training raises the score from 23.8 to 46.4, a gain of 22.6 points. This is the strongest listed result among open-weight models below 200B parameters. RepoHorizen also surpasses GPT-5.5 and Gemini-3.1 Pro, which score 45.1 and 33.4, respectively, as well as larger open-weight models DeepSeek-V4-Pro and MiniMax-M3, which score 38.5 and 42.1. Improvements extend to all four remaining benchmarks, with gains of 1.3 to 6.8 points across issue resolution and cross-domain long-horizon tasks in Terminal-Bench 2.1. These results suggest that RepoFactory training transfers beyond repository generation to broader software engineering and terminal work at fixed model scale. Environment evolution further strengthens this performance across all five benchmarks, adding 5.1 points on NL2Repo and 2.1 points on average over the no-evolution variant. These additional gains support the effectiveness of our proxy metrics in guiding environment evolution that improve training value.


\subsection{Environment Scaling}

\begin{figure}[h]
  \centering
  \includegraphics[width=0.45\textwidth]{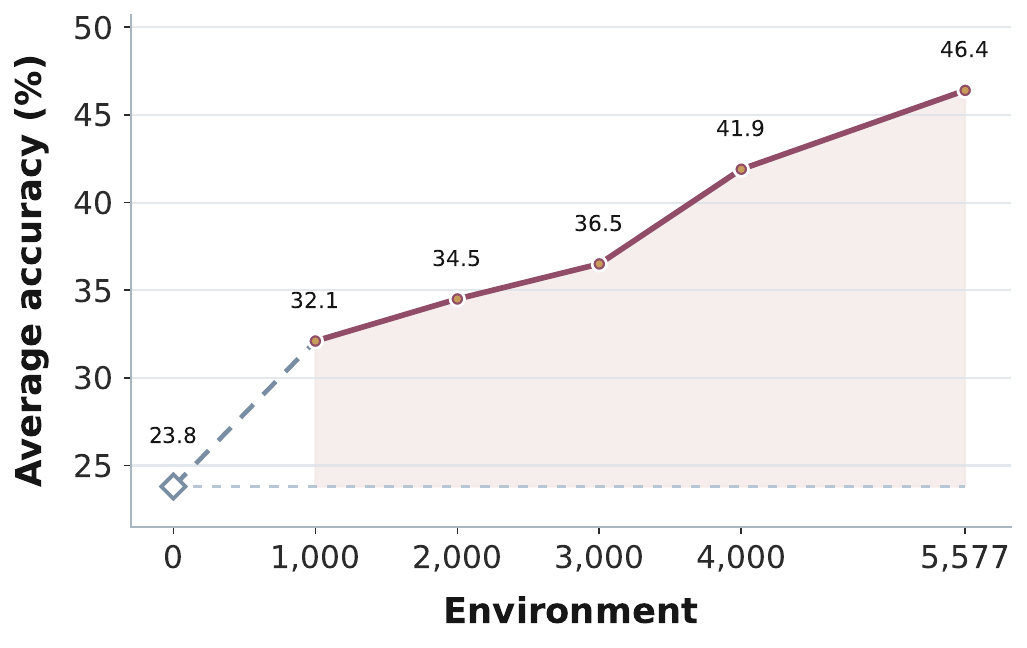}
\caption{Average accuracy as a function of the number of RepoFactory training environments.}
  \label{fig:env-scale}
\end{figure}

Figure~\ref{fig:env-scale} shows how NL2Repo performance changes as the training set grows. Performance increases steadily from 23.8\% without RepoFactory environments to 46.4\% with 5,577 environments, consistent with the 22.6-point gain reported in Table~\ref{tab:main-results}. The improvement continues at the largest scale tested, with no evidence of saturation.

The gains become more pronounced beyond 3,000 environments. Adding the remaining 2,577 environments raises performance from 36.5\% to 46.4\%, suggesting that the larger collection exposes the model to repository structures and behavioral requirements that smaller subsets do not cover. Scaling the environment set therefore expands the supervision available for long-horizon repository reconstruction and continues to improve performance well beyond the earliest additions.

\subsection{Scaling Across Model Sizes}

\begin{figure}[h]
  \centering
  \includegraphics[width=0.45\textwidth]{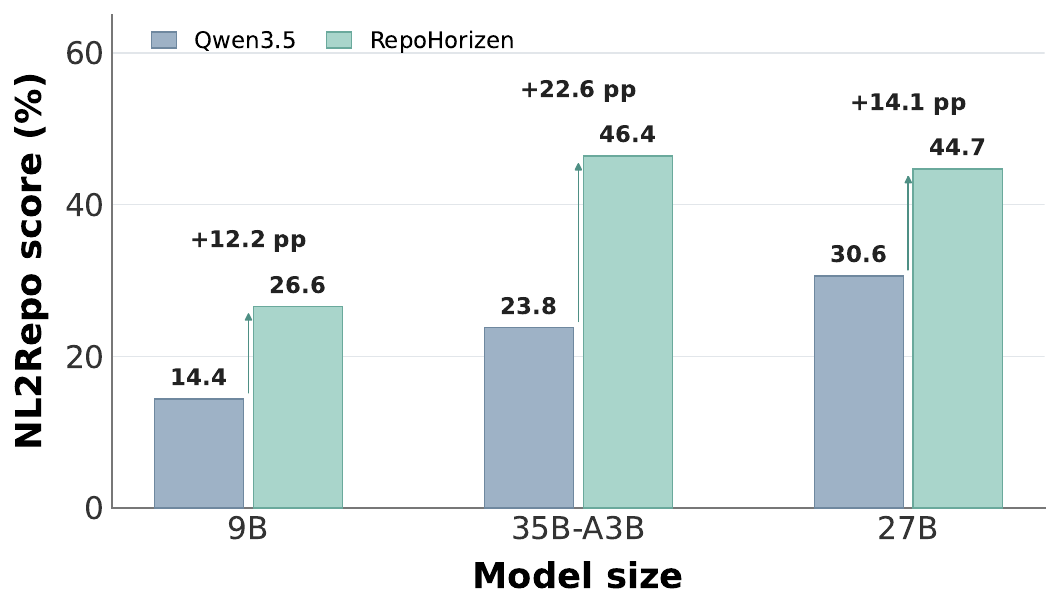}
\caption{NL2Repo performance across model sizes.}
  \label{fig:cross_model}
\end{figure}

We examine how the benefits of RepoFactory training vary with model capacity. Figure~\ref{fig:cross_model} reports NL2Repo gains of 12.2 to 22.6 percentage points across the 9B, 27B, and 35B-A3B models, showing that full-repository task remains useful across different parameter budgets. The 35B-A3B model gains the most despite the 27B model's higher baseline, suggesting that stronger initial performance does not necessarily translate into greater benefit from this training.



\begin{figure*}[t]
  \centering
  \includegraphics[width=0.98\textwidth]{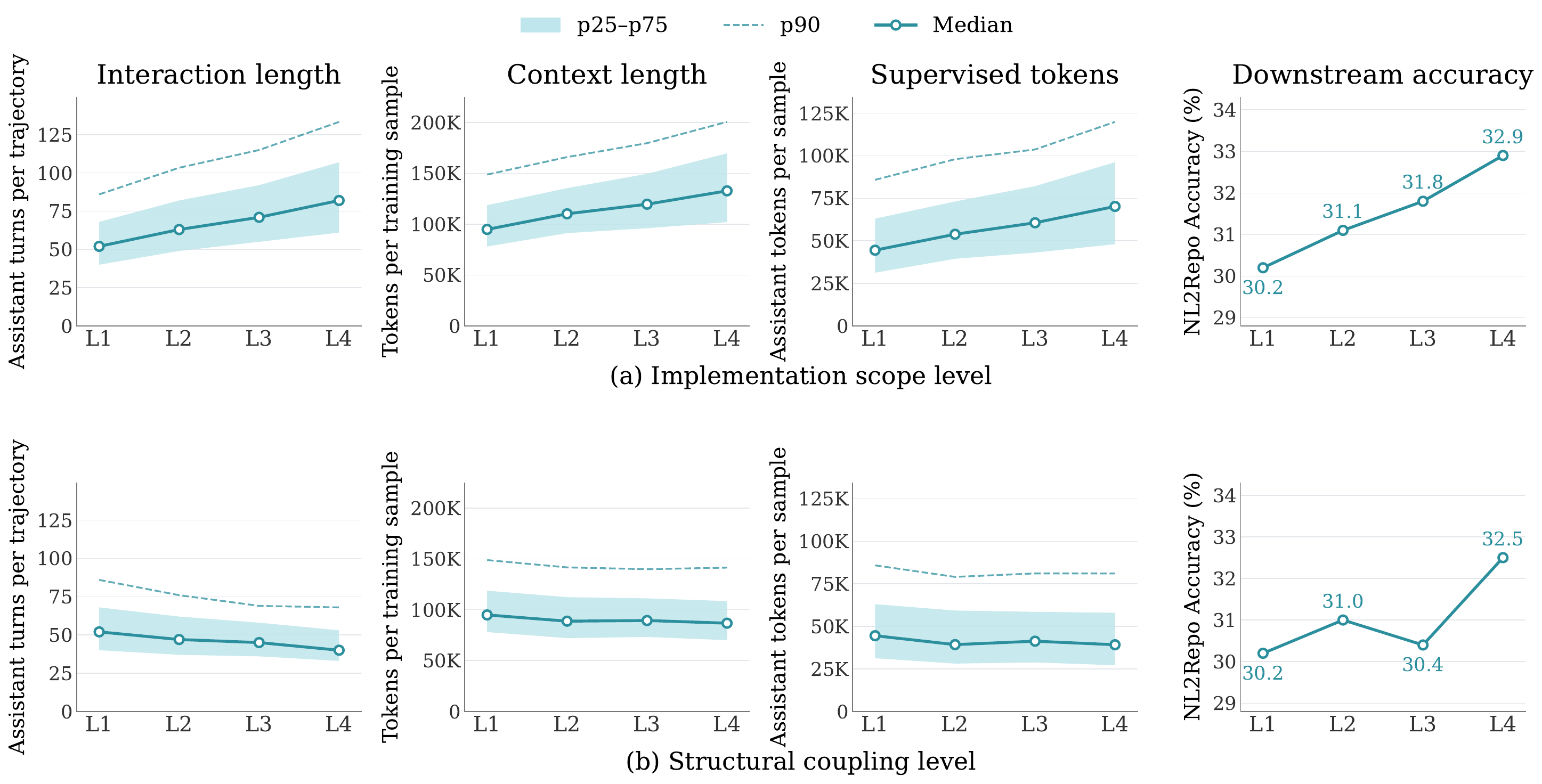}
  \caption{Statistics of the four difficulty levels and the corresponding NL2Repo results.}
  \label{fig:level-overview}
\end{figure*}




\begin{figure}[t]
  \centering
  \includegraphics[width=0.45\textwidth]{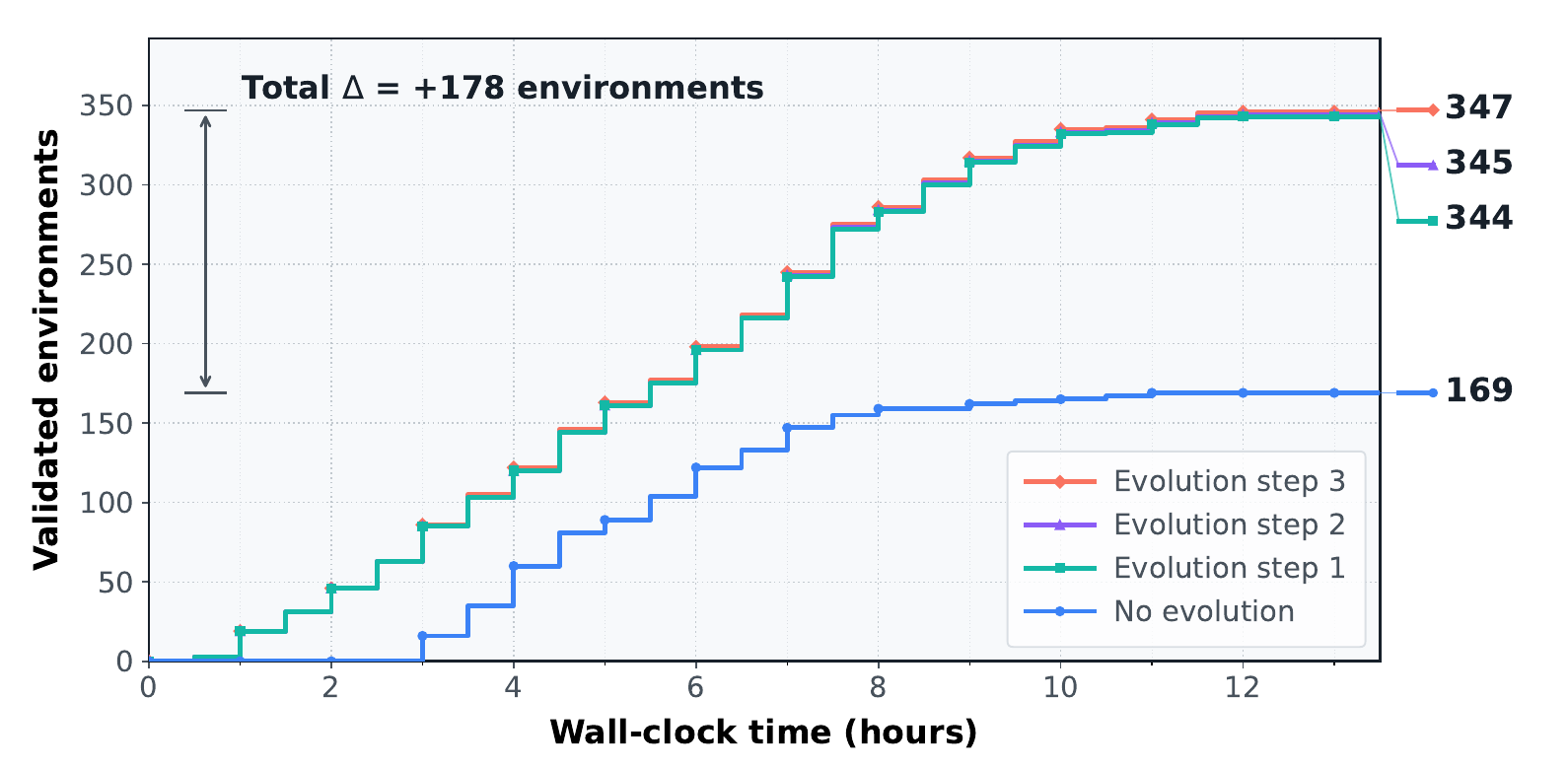}
  \caption{Cumulative number of validated environments over wall-clock time, sampled every 30 minutes.}
  \label{fig:eel}
\end{figure}

\subsection{Effect of the Data Source}

\begin{table}[t]
\centering
\footnotesize
\setlength{\tabcolsep}{3pt}
\begin{tabular}{lcccccc}
\toprule
Training Data & Open & N2R & Ver. & Pro & ML & TB \\
\midrule
\multicolumn{7}{l}{\textit{Reported by prior work (varying harnesses)}} \\
DeNovoSWE      & \textcolor{BrickRed}{\ding{55}} & 27.1 & -- & -- & -- & -- \\
MindForge      & \textcolor{BrickRed}{\ding{55}} & 23.5 & 73.8 & 51.3 & 67.8 & -- \\
Refact-92B     & \textcolor{BrickRed}{\ding{55}} & 28.9 & -- & -- & -- & -- \\
TrajLong       & \textcolor{BrickRed}{\ding{55}} & 13.8 & 51.1 & 34.3 & -- & 12.4 \\
\midrule
\multicolumn{7}{l}{\textit{Ours (Claude Code harness)}} \\
OpenSWE        & \textcolor{ForestGreen}{\checkmark} & 25.2 & 64.6 & 40.5 & 57.0 & 37.1 \\
R2E            & \textcolor{ForestGreen}{\checkmark} & 22.6 & 62.0 & 40.5 & 59.7 & 29.2 \\
ScaleSWE       & \textcolor{ForestGreen}{\checkmark} & 24.0 & 56.2 & 40.9 & 51.3 & 29.2 \\
SWE-smith      & \textcolor{ForestGreen}{\checkmark} & 22.6 & 59.4 & 39.0 & 55.3 & 31.5 \\
SWE-rebench    & \textcolor{ForestGreen}{\checkmark} & 23.1 & 68.8 & 42.0 & 59.0 & 29.2 \\
SWE-rebench-v2 & \textcolor{ForestGreen}{\checkmark} & 24.0 & 62.4 & 36.8 & 54.0 & 33.7 \\
\rowcolor[HTML]{DFE7F5}
RepoFactory & \textcolor{ForestGreen}{\checkmark} & \textbf{46.4} & \textbf{72.2} & \textbf{47.6} & \textbf{61.0} & \textbf{40.5} \\
\bottomrule
\end{tabular}
\caption{Performance of models trained on different SWE data sources, in percent.
N2R = NL2Repo, Ver.\ = SWE-bench Verified, Pro = SWE-bench Pro,
ML = SWE-bench Multilingual, TB = Terminal-Bench 2.1.
Best result among our runs in \textbf{bold}.}
\label{tab:data-comparison}
\end{table}

We compare RepoFactory with existing SWE datasets to assess the training value of its environments. We collect trajectories on six open datasets and apply the same cleaning procedure as for RepoFactory. Each baseline contains 11,000 training trajectories, slightly more than RepoFactory. We fine-tune Qwen3.5-35B-A3B separately on each collection and evaluate the models with the Claude Code harness. Table~\ref{tab:data-comparison} includes both published results and these controlled comparisons. Among the published results, obtained with varying harnesses, the highest NL2Repo score is 28.9. RepoFactory reaches 46.4 and leads all six open-data baselines on all five benchmarks. On NL2Repo, its advantage over the strongest baseline, OpenSWE, is 21.2 points. The teacher is the same across our runs, so its strength alone cannot explain this advantage. The results support the value of the training experience provided by RepoFactory's environments.

\subsection{Ablation of the Evolution Objective}
\label{sec:evolution-analysis}

The evolution objective must support both the recovery of usable environments and improvements in training value. We evaluate these effects separately, using oracle validation to measure candidate recovery and downstream training to assess $S$ and $C$ as optimization targets. For recovery, we uniformly sample 500 repositories from the filtered candidate set and compare runs allowing up to three repair steps and none. We compare the number of environments satisfying $A=1$, with concurrency, timeouts, and infrastructure retry budgets held fixed.

Figure~\ref{fig:eel} shows that evolution increases the number of validated environments from 169 to 347, raising yield from 33.8\% to 69.4\% under the same acceptance criterion. These gains suggest that many initial failures stem from construction defects that can be corrected through execution feedback. Most of the improvement comes from the first repair, with smaller gains from subsequent steps. Recovering these otherwise discarded candidates expands the supply of valid environments from the same repository pool, supporting environment generation at scale.

To assess whether optimizing $S$ and $C$ improves training value, we run two independent evolution experiments from the same 1,000 randomly sampled environments. Each experiment targets one metric while retaining validity checks. We collect environments after rounds 3, 6, 9, and 12 as Levels~1 through~4, respectively. Each level supplies 2,000 trajectories for fine-tuning Qwen3.5-35B-A3B with the same initialization and optimization settings.

As shown in Figure~\ref{fig:level-overview}, optimizing $S$ produces longer interactions and contexts and more supervised tokens. NL2Repo accuracy increases steadily from 30.2\% at Level~1 to 32.9\% at Level~4. Expanding implementation scope thus provides more extensive development experience within a fixed number of trajectories, accompanied by improved reconstruction performance.

Optimizing $C$ produces a different pattern. Accuracy fluctuates at intermediate levels but reaches 32.5\% at Level~4, compared with 30.2\% at Level~1. Interaction length decreases, while context length and supervised token counts decline only slightly. The improvement therefore does not depend on longer trajectories or additional supervision. This complements the scope ablation, suggesting that training value depends not only on the extent of implementation work but also on the relationships among the components being implemented.

\section{Conclusion}

We presented RepoFactory, an automated pipeline that constructs verifiable environments for full-repository generation from real-world projects. Its Environment Evolution Loop uses timely, low-cost feedback to optimize validity, implementation scope, and structural coupling, recovering failed conversions and improving task value. The resulting 10,159 environments span 17 software domains. Training Qwen3.5-35B-A3B on these environments yields RepoHorizen-35B-A3B, with gains of 22.6 percentage points on NL2Repo-Bench and 8.5 points on average across five coding and agentic benchmarks. Environment evolution contributes a further 5.1 and 2.1 points over training without evolution. These findings highlight the value of repository-level environments for coding-agent training and support low-cost proxy objectives as a practical means of improving their training value at scale.

\section{Limitations}

RepoFactory currently focuses on Python projects, with relatively limited coverage of other programming languages. Our proxy metrics provide useful, low-cost guidance for environment evolution, but characterize task content at a broad structural level. They do not yet explicitly distinguish the capabilities required by individual tasks. Extending language coverage and incorporating more detailed information about these capability requirements could support a wider range of development settings and more targeted training.

\bibliography{main}

@String(ICLR = {Int. Conf. Learn. Represent.})

@String(ICLR  = {ICLR})

@article{xu2026autolab,
  title={AutoLab: Can Frontier Models Solve Long-Horizon Auto Research and Engineering Tasks?},
  author={Xu, Zhangchen and Chen, Junda and Huang, Yue and Jiang, Dongfu and Chen, Jiefeng and Hua, Hang and Wu, Zijian and Liu, Zheyuan and He, Zexue and Li, Lichi and others},
  journal={arXiv preprint arXiv:2606.05080},
  year={2026}
}

@article{deepseek2606deepseek,
  title={Deepseek-v4: Towards highly efficient million-token context intelligence, 2026},
  author={DeepSeek-AI, Anyi Xu and Lin, Bangcai and Xue, Bing and Wang, Bingxuan and Xu, Bingzheng and Wu, Bochao and Zhang, Bowei and Lin, Chaofan and Dong, Chen and Ling, Chenchen and others},
  journal={URL https://arxiv. org/abs/2606.19348}
}

@article{li2026recursive,
  title={Recursive synthesis for long-horizon terminal tasks},
  author={Li, Zhongzhi and Shi, Yucheng and Li, Zongxia and Wang, Ruhan and Li, Anhao and Huang, Zixun and Yang, Junyao and Ke, Lei and Liu, Ninghao and Mi, Haitao and others},
  journal={arXiv preprint arXiv:2608.05466},
  year={2026}
}

@article{bajoria2026language,
  title={From language models to agentic ai: A survey of autonomous, action-enabled, and collaborative llm agents},
  author={Bajoria, Sparsh and Ranjan, Shreyanshu and M, Adhitya and Suri, Vineet and Hassija, Vikas and Chamola, Vinay and Hussain, Amir},
  journal={Cognitive Computation},
  volume={18},
  number={1},
  pages={103},
  year={2026},
  publisher={Springer}
}

@article{tian2026beyond,
  title={Beyond function calling: Benchmarking tool-using agents under tool-environment unreliability},
  author={Tian, Yang and Shi, Zhengpeng and Zhou, Yu and Zhao, Bo},
  journal={arXiv preprint arXiv:2606.25819},
  year={2026}
}

@article{icoder,
  author       = {Jian Yang and
                  Wei Zhang and
                  Jiajun Wu and
                  Junhang Cheng and
                  Shawn Guo and
                  Haowen Wang and
                  Weicheng Gu and
                  Yaxin Du and
                  Joseph Li and
                  Fanglin Xu and
                  Yizhi Li and
                  Lin Jing and
                  Yuanbo Wang and
                  Yuhan Gao and
                  Ruihao Gong and
                  Chuan Hao and
                  Ran Tao and
                  Aishan Liu and
                  Tuney Zheng and
                  Ganqu Cui and
                  Zhoujun Li and
                  Mingjie Tang and
                  Chenghua Lin and
                  Wayne Xin Zhao and
                  Xianglong Liu and
                  Ming Zhou and
                  Bryan Dai and
                  Weifeng Lv},
  title        = {InCoder-32B: Code Foundation Model for Industrial Scenarios},
  journal      = {CoRR},
  volume       = {abs/2603.16790},
  year         = {2026},
  url          = {https://doi.org/10.48550/arXiv.2603.16790},
  doi          = {10.48550/ARXIV.2603.16790},
  eprinttype   = {arXiv},
  eprint       = {2603.16790},
  bibsource    = {dblp computer science bibliography, https://dblp.org}
}

@article{team2026kimi,
  title={Kimi k3: Open frontier intelligence},
  author={Team, Kimi and Bai, Tongtong and Bai, Yifan and Bao, Yiping and Cai, Jianfeng and Cai, Xinyuan and Cao, Peizhou and Cao, Yuxuan and Chai, Ziwei and Charles, Y and others},
  journal={arXiv preprint arXiv:2607.24653},
  year={2026}
}

@inproceedings{EnvScaler,
  author       = {Xiaoshuai Song and
                  Haofei Chang and
                  Guanting Dong and
                  Yutao Zhu and
                  Ji{-}Rong Wen and
                  Zhicheng Dou},
  editor       = {Maria Liakata and
                  Viviane P. Moreira and
                  Jiajun Zhang and
                  David Jurgens},
  title        = {EnvScaler: Scaling Tool-Interactive Environments for {LLM} Agent via
                  Programmatic Synthesis},
  booktitle    = {Findings of the Association for Computational Linguistics, {ACL} 2026,
                  San Diego, California, United States, July 2-7, 2026},
  pages        = {8326--8357},
  publisher    = {Association for Computational Linguistics},
  year         = {2026},
  url          = {https://doi.org/10.18653/v1/2026.findings-acl.407},
  doi          = {10.18653/V1/2026.FINDINGS-ACL.407},
  bibsource    = {dblp computer science bibliography, https://dblp.org}
}

@article{ScaleEnv,
  author       = {Dunwei Tu and
                  Hongyan Hao and
                  Hansi Yang and
                  Yihao Chen and
                  Yi{-}Kai Zhang and
                  Zhikang Xia and
                  Yu Yang and
                  Yueqing Sun and
                  Xingchen Liu and
                  Furao Shen and
                  Qi Gu and
                  Hui Su and
                  Xunliang Cai},
  title        = {ScaleEnv: Scaling Environment Synthesis from Scratch for Generalist
                  Interactive Tool-Use Agent Training},
  journal      = {CoRR},
  volume       = {abs/2602.06820},
  year         = {2026},
  url          = {https://doi.org/10.48550/arXiv.2602.06820},
  doi          = {10.48550/ARXIV.2602.06820},
  eprinttype   = {arXiv},
  eprint       = {2602.06820},
  bibsource    = {dblp computer science bibliography, https://dblp.org}
}

@article{glm2026glm,
  title={GLM-5: from Vibe Coding to Agentic Engineering. arXiv preprint (2026)},
  author={GLM Team and others},
  journal={arXiv preprint arXiv:2602.15763},
  year={2026}
}

@article{Autodata,
  author       = {Ilia Kulikov and
                  Chenxi Whitehouse and
                  Tianhao Wu and
                  Yixin Nie and
                  Swarnadeep Saha and
                  Eryk Helenowski and
                  Weizhe Yuan and
                  Olga Golovneva and
                  Jack Lanchantin and
                  Yoram Bachrach and
                  Jakob N. Foerster and
                  Xian Li and
                  Han Fang and
                  Sainbayar Sukhbaatar and
                  Jason Weston},
  title        = {Autodata: An agentic data scientist to create high quality synthetic
                  data},
  journal      = {CoRR},
  volume       = {abs/2606.25996},
  year         = {2026},
  url          = {https://doi.org/10.48550/arXiv.2606.25996},
  doi          = {10.48550/ARXIV.2606.25996},
  eprinttype   = {arXiv},
  eprint       = {2606.25996},
  bibsource    = {dblp computer science bibliography, https://dblp.org}
}

@article{Framework,
  author       = {Jincheng Ren and
                  Siwei Wu and
                  Yizhi Li and
                  Kang Zhu and
                  Shu Xu and
                  Boyu Feng and
                  Ruibin Yuan and
                  Wei Zhang and
                  Riza Batista{-}Navarro and
                  Jian Yang and
                  Chenghua Lin},
  title        = {A Self-Evolving Framework for Efficient Terminal Agents via Observational
                  Context Compression},
  journal      = {CoRR},
  volume       = {abs/2604.19572},
  year         = {2026},
  url          = {https://doi.org/10.48550/arXiv.2604.19572},
  doi          = {10.48550/ARXIV.2604.19572},
  eprinttype   = {arXiv},
  eprint       = {2604.19572},
  bibsource    = {dblp computer science bibliography, https://dblp.org}
}

@article{Training,
  author       = {Yiqi Zhu and
                  Apurva Gandhi and
                  Graham Neubig},
  title        = {Training Versatile Coding Agents in Synthetic Environments},
  journal      = {CoRR},
  volume       = {abs/2512.12216},
  year         = {2025},
  url          = {https://doi.org/10.48550/arXiv.2512.12216},
  doi          = {10.48550/ARXIV.2512.12216},
  eprinttype   = {arXiv},
  eprint       = {2512.12216},
  bibsource    = {dblp computer science bibliography, https://dblp.org}
}

@article{Endless,
  author       = {Kanishk Gandhi and
                  Shivam Garg and
                  Noah D. Goodman and
                  Dimitris Papailiopoulos},
  title        = {Endless Terminals: Scaling {RL} Environments for Terminal Agents},
  journal      = {CoRR},
  volume       = {abs/2601.16443},
  year         = {2026},
  url          = {https://doi.org/10.48550/arXiv.2601.16443},
  doi          = {10.48550/ARXIV.2601.16443},
  eprinttype   = {arXiv},
  eprint       = {2601.16443},
  bibsource    = {dblp computer science bibliography, https://dblp.org}
}

@article{LiteCoder-Terminal,
  author       = {Xiaoxuan Peng and
                  Kaiqi Zhang and
                  Xinyu Lu and
                  Boxi Cao and
                  Yaojie Lu and
                  Hongyu Lin and
                  Xianpei Han and
                  Le Sun},
  title        = {LiteCoder-Terminal: Scaling Long-Horizon Terminal Environments for
                  Learning Language Agents},
  journal      = {CoRR},
  volume       = {abs/2605.29559},
  year         = {2026},
  url          = {https://doi.org/10.48550/arXiv.2605.29559},
  doi          = {10.48550/ARXIV.2605.29559},
  eprinttype   = {arXiv},
  eprint       = {2605.29559},
  bibsource    = {dblp computer science bibliography, https://dblp.org}
}

@article{CLI-Universe,
  author       = {Zhanbo Hua and
                  Yifan Yao and
                  Weihao Xie and
                  Yongchi Zhao and
                  Minghao Liu and
                  Ruizhi Qiu and
                  Zhewei Huang and
                  Zun Wang and
                  Yiyan Ji and
                  Yunhai Ye and
                  Letian Zhu and
                  Xinping Lei and
                  Han Li and
                  Zhiyuan Ma and
                  Zili Wang and
                  Zhaoxiang Zhang and
                  Jiaheng Liu},
  title        = {CLI-Universe: Towards Verifiable Task Synthesis Engine for Terminal
                  Agents},
  journal      = {CoRR},
  volume       = {abs/2606.22883},
  year         = {2026},
  url          = {https://doi.org/10.48550/arXiv.2606.22883},
  doi          = {10.48550/ARXIV.2606.22883},
  eprinttype   = {arXiv},
  eprint       = {2606.22883},
  bibsource    = {dblp computer science bibliography, https://dblp.org}
}

@inproceedings{Self-Play,
  author       = {Zixiang Chen and
                  Yihe Deng and
                  Huizhuo Yuan and
                  Kaixuan Ji and
                  Quanquan Gu},
  editor       = {Ruslan Salakhutdinov and
                  Zico Kolter and
                  Katherine A. Heller and
                  Adrian Weller and
                  Nuria Oliver and
                  Jonathan Scarlett and
                  Felix Berkenkamp},
  title        = {Self-Play Fine-Tuning Converts Weak Language Models to Strong Language
                  Models},
  booktitle    = {Forty-first International Conference on Machine Learning, {ICML} 2024,
                  Vienna, Austria, July 21-27, 2024},
  series       = {Proceedings of Machine Learning Research},
  volume       = {235},
  pages        = {6621--6642},
  publisher    = {{PMLR} / OpenReview.net},
  year         = {2024},
  url          = {https://proceedings.mlr.press/v235/chen24j.html},
  bibsource    = {dblp computer science bibliography, https://dblp.org}
}

@article{MM-Zero,
  author       = {Zongxia Li and
                  Hongyang Du and
                  Chengsong Huang and
                  Xiyang Wu and
                  Lantao Yu and
                  Yicheng He and
                  Jing Xie and
                  Xiaomin Wu and
                  Zhichao Liu and
                  Jiarui Zhang and
                  Fuxiao Liu},
  title        = {MM-Zero: Self-Evolving Multi-Model Vision Language Models From Zero
                  Data},
  journal      = {CoRR},
  volume       = {abs/2603.09206},
  year         = {2026},
  url          = {https://doi.org/10.48550/arXiv.2603.09206},
  doi          = {10.48550/ARXIV.2603.09206},
  eprinttype   = {arXiv},
  eprint       = {2603.09206},
  bibsource    = {dblp computer science bibliography, https://dblp.org}
}

@article{BenchEvolver,
  author       = {Yangzhen Wu and
                  Aaron J. Li and
                  Wenjie Ma and
                  Li Cao and
                  Ziheng Zhou and
                  Mert Cemri and
                  Shu Liu and
                  Yuran Xiu and
                  Chenxiao Yan and
                  Haikun Zhao and
                  Bin Yu and
                  Ion Stoica and
                  Dawn Song},
  title        = {BenchEvolver: Frontier Task Synthesis via Solution-Centric Evolution},
  journal      = {CoRR},
  volume       = {abs/2606.01286},
  year         = {2026},
  url          = {https://doi.org/10.48550/arXiv.2606.01286},
  doi          = {10.48550/ARXIV.2606.01286},
  eprinttype   = {arXiv},
  eprint       = {2606.01286},
  bibsource    = {dblp computer science bibliography, https://dblp.org}
}

@article{seta,
  author       = {Qijia Shen and
                  Zhiqi Huang and
                  Vamsidhar Kamanuru and
                  Aznaur Aliev and
                  Jay Rainton and
                  Ahmed Awelkair and
                  Zhichen Zeng and
                  Jiajun Li and
                  Shi Dong and
                  Yueming Yuan and
                  Boyuan Ma and
                  Qizheng Zhang and
                  Jiwei Fu and
                  Yuzhen Mao and
                  Wendong Fan and
                  Ping Nie and
                  Philip Torr and
                  Bernard Ghanem and
                  Changran Hu and
                  Jonathan Lingjie Li and
                  Urmish Thakker and
                  Guohao Li},
  title        = {{SETA:} Scaling Environments for Terminal Agents},
  journal      = {CoRR},
  volume       = {abs/2607.10891},
  year         = {2026},
  url          = {https://doi.org/10.48550/arXiv.2607.10891},
  doi          = {10.48550/ARXIV.2607.10891},
  eprinttype   = {arXiv},
  eprint       = {2607.10891},
  bibsource    = {dblp computer science bibliography, https://dblp.org}
}

@article{NL2Repo-Bench,
  author       = {Jingzhe Ding and
                  Shengda Long and
                  Changxin Pu and
                  Huan Zhou and
                  Hongwan Gao and
                  Xiang Gao and
                  Chao He and
                  Yue Hou and
                  Fei Hu and
                  Zhaojian Li and
                  Weiran Shi and
                  Zaiyuan Wang and
                  Daoguang Zan and
                  Chenchen Zhang and
                  Xiaoxu Zhang and
                  Qizhi Chen and
                  Xianfu Cheng and
                  Bo Deng and
                  Qingshui Gu and
                  Kai Hua and
                  Juntao Lin and
                  Pai Liu and
                  Mingchen Li and
                  Xuanguang Pan and
                  Zifan Peng and
                  Yujia Qin and
                  Yong Shan and
                  Zhewen Tan and
                  Weihao Xie and
                  Zihan Wang and
                  Yishuo Yuan and
                  Jiayu Zhang and
                  Enduo Zhao and
                  Yunfei Zhao and
                  He Zhu and
                  Chenyang Zou and
                  Ming Ding and
                  Jianpeng Jiao and
                  Jiaheng Liu and
                  Minghao Liu and
                  Qian Liu and
                  Chongyao Tao and
                  Jian Yang and
                  Tong Yang and
                  Zhaoxiang Zhang and
                  Xinjie Chen and
                  Wenhao Huang and
                  Ge Zhang},
  title        = {NL2Repo-Bench: Towards Long-Horizon Repository Generation Evaluation
                  of Coding Agents},
  journal      = {CoRR},
  volume       = {abs/2512.12730},
  year         = {2025},
  url          = {https://doi.org/10.48550/arXiv.2512.12730},
  doi          = {10.48550/ARXIV.2512.12730},
  eprinttype   = {arXiv},
  eprint       = {2512.12730},
  bibsource    = {dblp computer science bibliography, https://dblp.org}
}

@article{SWE-Bench-pro,
  author       = {Xiang Deng and
                  Jeff Da and
                  Edwin Pan and
                  Yannis Yiming He and
                  Charles Ide and
                  Kanak Garg and
                  Niklas Lauffer and
                  Andrew Park and
                  Nitin Pasari and
                  Chetan Rane and
                  Karmini Sampath and
                  Maya Krishnan and
                  Srivatsa Kundurthy and
                  Sean Hendryx and
                  Zifan Wang and
                  Chen Bo Calvin Zhang and
                  Noah Jacobson and
                  Bing Liu and
                  Brad Kenstler},
  title        = {SWE-Bench Pro: Can {AI} Agents Solve Long-Horizon Software Engineering
                  Tasks?},
  journal      = {CoRR},
  volume       = {abs/2509.16941},
  year         = {2025},
  url          = {https://doi.org/10.48550/arXiv.2509.16941},
  doi          = {10.48550/ARXIV.2509.16941},
  eprinttype   = {arXiv},
  eprint       = {2509.16941},
  bibsource    = {dblp computer science bibliography, https://dblp.org}
}

@inproceedings{Multi-SWE-bench,
  author       = {Daoguang Zan and
                  Zhirong Huang and
                  Wei Liu and
                  Hanwu Chen and
                  Shulin Xin and
                  Linhao Zhang and
                  Qi Liu and
                  Aoyan Li and
                  Lu Chen and
                  Xiaojian Zhong and
                  Siyao Liu and
                  Yongsheng Xiao and
                  Liangqiang Chen and
                  Yuyu Zhang and
                  Jing Su and
                  Tianyu Liu and
                  Rui Long and
                  Ming Ding and
                  Liang Xiang},
  editor       = {Danielle Belgrave and
                  Cheng Zhang and
                  Laura N. Montoya and
                  Hsuan{-}Tien Lin and
                  Razvan Pascanu and
                  Piotr Koniusz and
                  Marzyeh Ghassemi and
                  Nancy Chen and
                  Iv{\'{a}}n Vladimir Meza Ru{\'{\i}}z and
                  Arturo Loaiza{-}Bonilla},
  title        = {Multi-SWE-bench: {A} Multilingual Benchmark for Issue Resolving},
  booktitle    = {Advances in Neural Information Processing Systems 38: Annual Conference
                  on Neural Information Processing Systems 2025, NeurIPS 2025, San Diego,
                  CA, USA, December 2-7, 2025 / Mexico City, Mexico, November 30 - December
                  5, 2025},
  year         = {2025},
  url          = {http://papers.nips.cc/paper\_files/paper/2025/hash/5afa9cb1e917b898ad418216dc726fbd-Abstract-Datasets\_and\_Benchmarks\_Track.html},
  bibsource    = {dblp computer science bibliography, https://dblp.org}
}

@inproceedings{Terminal-bench,
  title={Terminal-bench: Benchmarking agents on hard, realistic tasks in command line interfaces},
  author={Merrill, Mike and Shaw, Alexander and Carlini, Nicholas and Li, Boxuan and Raj, Harsh and Bercovich, Ivan and Shi, Lin and Shin, Jeong and Walshe, Thomas and Buchanan, E Kelly and others},
  booktitle={International Conference on Learning Representations},
  volume={2026},
  pages={40903--40986},
  year={2026}
}

@inproceedings{SWE-bench,
  author       = {Carlos E. Jimenez and
                  John Yang and
                  Alexander Wettig and
                  Shunyu Yao and
                  Kexin Pei and
                  Ofir Press and
                  Karthik R. Narasimhan},
  title        = {SWE-bench: Can Language Models Resolve Real-world Github Issues?},
  booktitle    = {The Twelfth International Conference on Learning Representations,
                  {ICLR} 2024, Vienna, Austria, May 7-11, 2024},
  publisher    = {OpenReview.net},
  year         = {2024},
  url          = {https://openreview.net/forum?id=VTF8yNQM66},
  bibsource    = {dblp computer science bibliography, https://dblp.org}
}

@inproceedings{SWE-Gym,
  author       = {Jiayi Pan and
                  Xingyao Wang and
                  Graham Neubig and
                  Navdeep Jaitly and
                  Heng Ji and
                  Alane Suhr and
                  Yizhe Zhang},
  editor       = {Aarti Singh and
                  Maryam Fazel and
                  Daniel Hsu and
                  Simon Lacoste{-}Julien and
                  Felix Berkenkamp and
                  Tegan Maharaj and
                  Kiri Wagstaff and
                  Jerry Zhu},
  title        = {Training Software Engineering Agents and Verifiers with SWE-Gym},
  booktitle    = {Forty-second International Conference on Machine Learning, {ICML}
                  2025, Vancouver, BC, Canada, July 13-19, 2025},
  series       = {Proceedings of Machine Learning Research},
  volume       = {267},
  publisher    = {{PMLR} / OpenReview.net},
  year         = {2025},
  url          = {https://proceedings.mlr.press/v267/pan25g.html},
  bibsource    = {dblp computer science bibliography, https://dblp.org}
}

@inproceedings{SWE-smith,
  author       = {John Yang and
                  Kilian Lieret and
                  Carlos E. Jimenez and
                  Alexander Wettig and
                  Kabir Khandpur and
                  Yanzhe Zhang and
                  Binyuan Hui and
                  Ofir Press and
                  Ludwig Schmidt and
                  Diyi Yang},
  editor       = {Danielle Belgrave and
                  Cheng Zhang and
                  Laura N. Montoya and
                  Hsuan{-}Tien Lin and
                  Razvan Pascanu and
                  Piotr Koniusz and
                  Marzyeh Ghassemi and
                  Nancy Chen and
                  Iv{\'{a}}n Vladimir Meza Ru{\'{\i}}z and
                  Arturo Loaiza{-}Bonilla},
  title        = {SWE-smith: Scaling Data for Software Engineering Agents},
  booktitle    = {Advances in Neural Information Processing Systems 38: Annual Conference
                  on Neural Information Processing Systems 2025, NeurIPS 2025, San Diego,
                  CA, USA, December 2-7, 2025 / Mexico City, Mexico, November 30 - December
                  5, 2025},
  year         = {2025},
  url          = {http://papers.nips.cc/paper\_files/paper/2025/hash/8b86cf5ace600c48fd188efbb8dedec8-Abstract-Datasets\_and\_Benchmarks\_Track.html},
  bibsource    = {dblp computer science bibliography, https://dblp.org}
}

@techreport{team2026ai4ai,
  title={AI4AI at scale: A full-pipeline system for enhancing LLM agentic capabilities},
  author={Team, XYZ Agentic},
  year={2026},
  institution={Technical report, XYZ AI Lab, July 2026. URL https://xyz-lab. ai/blogs/ai4ai~…}
}

@article{yang2026iquest,
  title={IQuest-Q1: Advancing Foundation Capabilities for Agentic CLI Systems},
  author={Yang, A Jian and Huang, An and Dan, Bo and Mu, Chen and Yang, Chengran and Shi, Chuan and Chang, Feng and Li, Hongxin and Tao, Hongyuan and Zhang, Jinxing and others},
  journal={arXiv preprint arXiv:2610.30949},
  year={2026}
}

@article{zhao2026denovoswe,
  title={DeNovoSWE: Scaling Long-Horizon Environments for Generating Entire Repositories from Scratch},
  author={Zhao, Jiale and Chen, Guoxin and Meng, Fanzhe and Zhao, Wayne Xin and Song, Ruihua and Wen, Ji-Rong and Jia, Kai},
  journal={arXiv preprint arXiv:2606.10728},
  year={2026}
}

@article{chen2026mindforge,
  title={MindForge: Teaching Small Language Models Whole-Life-Cycle Software Engineering via Source-Free Program Synthesis},
  author={Chen, Yihao and Chang, Shi and Chawa, Khaled and Lin, Feng and Chen, Boyuan and Wang, Shaowei and Hassan, Ahmed E},
  journal={arXiv preprint arXiv:2607.27146},
  year={2026}
}

@article{yang2026programgym,
  title={ProgramGym: Scaling Training Environments for Long-Horizon Program Reproduction Agents},
  author={Yang, Chengyi and Bao, Renxu and Hou, Jue and Teng, Zongpei and Wang, Ziheng and Lin, Yujie and Huang, Chengsong and Zhu, Junnan and Qiu, Delai and Wang, Yining and others},
  journal={arXiv preprint arXiv:2610.13395},
  year={2026}
}

@article{DataMaster,
  author       = {Yaxin Du and
                  Xiyuan Yang and
                  Zhifan Zhou and
                  Wanxu Liu and
                  Zixing Lei and
                  Zimeng Chen and
                  Fenyi Liu and
                  Haotian Wu and
                  Yuzhu Cai and
                  Zexi Liu and
                  Xinyu Zhu and
                  Wenhao Wang and
                  Linfeng Zhang and
                  Chen Qian and
                  Siheng Chen},
  title        = {DataMaster: Data-Centric Autonomous {AI} Research},
  journal      = {CoRR},
  volume       = {abs/2605.10906},
  year         = {2026},
  url          = {https://doi.org/10.48550/arXiv.2605.10906},
  doi          = {10.48550/ARXIV.2605.10906},
  eprinttype   = {arXiv},
  eprint       = {2605.10906},
  bibsource    = {dblp computer science bibliography, https://dblp.org}
}

@article{ebrahimi2026software,
  title={Software Engineering LLMs: Why We Need Them and How Software Engineering Helps Build Them},
  author={Ebrahimi, Amir M and Hasan, Mohammed Mehedi and Daneshvar, Seyed Shayan and Lin, Jiahuei and Rajbahadur, Gopi Krishnan and Hassan, Ahmed E},
  journal={arXiv preprint arXiv:2610.33851},
  year={2026}
}

@article{peng2026trajlong,
  title={TrajLong: Co-Designing Agentic and Long-Context Supervision for Mid-Training},
  author={Peng, Miao and Zhang, Qintong and Chen, Nuo and Li, Yuhan and Yan, Guochen and Gu, Xinran and Wu, Hongqiu and Wang, Hai and Huang, Lydell and Zhang, Wentao and others},
  journal={arXiv preprint arXiv:2610.04973},
  year={2026}
}

\appendix

\clearpage
\section{Implementation Details}
\label{app:implementation-details}

This appendix describes the trajectory preparation and supervised fine-tuning configuration used for Qwen3.5-35B-A3B.

\subsection{Trajectory Filtering and Cleaning}
\label{app:trajectory-cleaning}

Before fine-tuning, we retain trajectories with rewards above $0.5$ and apply offline quality checks. Deterministic checks examine record validity, dialogue structure, consistency between tool calls and results, response completeness, and the formatting of reasoning and code. They also screen for repetition, interaction loops, contamination, potentially sensitive or harmful content, and attempts to obtain information outside the intended task environment. We discard trajectories with clear violations and flag ambiguous cases for review.

A language-model judge reviews the flagged trajectories. These trajectories are included in the SFT data only when the judge determines that the apparent violation is benign. Confirmed violations and unresolved cases remain excluded.

\subsection{SFT Hyperparameters}
\label{app:sft-config}

\begin{table}[h]
  \centering
  \begin{minipage}{0.78\linewidth}
  \setlength{\belowcaptionskip}{6pt}
  \begingroup
  \fontsize{9}{11}\selectfont
  \setlength{\tabcolsep}{6pt}
  \renewcommand{\arraystretch}{1.12}
  \setlength{\heavyrulewidth}{0.7pt}
  \setlength{\lightrulewidth}{0.35pt}
  \arrayrulecolor{black!60}
  \begin{tabularx}{\linewidth}{
    >{\raggedright\arraybackslash}X
    >{\raggedright\arraybackslash}X}
    \toprule
    \rowcolor{nblue!6}
    \textbf{Hyperparameter} & \textbf{Value} \\
    \midrule
    \rowcolor{black!2.5}
    Number of epochs & 3 \\
    Rollout / global batch size & 64 / 64 \\
    \rowcolor{black!2.5}
    Per-device micro-batch size & 1 \\
    Dynamic token batching & Enabled \\
    \rowcolor{black!2.5}
    Token budget per GPU & 80,000 \\
    Optimizer & AdamW \\
    \rowcolor{black!2.5}
    Peak / minimum learning rate & $1.0\times10^{-6}$ / $1.0\times10^{-7}$ \\
    Learning-rate schedule / warmup & Cosine / 10\% \\
    \rowcolor{black!2.5}
    Weight decay & 0.1 \\
    Adam coefficients & $\beta_1=0.9$, $\beta_2=0.98$ \\
    \midrule
    \rowcolor{black!2.5}
    Tensor parallel size & 2 \\
    Pipeline parallel size & 1 \\
    \rowcolor{black!2.5}
    Context parallel size & 4 \\
    Expert parallel size & 8 \\
    \rowcolor{black!2.5}
    Expert tensor parallel size & 1 \\
    Sequence parallelism & Enabled \\
    \bottomrule
  \end{tabularx}
  \endgroup
  \end{minipage}
  \caption{Supervised fine-tuning hyperparameters for Qwen3.5-35B-A3B.}
  \label{tab:sft-hyperparameters}
\end{table}

We fine-tune the model, using the settings in Table~\ref{tab:sft-hyperparameters}. The data are shuffled during training, with no online generation. Following the Qwen3.5 multi-turn objective, we compute the loss over assistant spans, including tool calls. System, user, and tool-result turns provide context but receive no loss.

The rollout and global batch sizes are both 64, and the per-device micro-batch size is 1. Dynamic token batching limits each GPU to 80,000 tokens, with no separate sequence-length cap. Tensor, context, and expert parallel sizes are 2, 4, and 8, respectively. Sequence parallelism is enabled, while pipeline and expert tensor parallel sizes are both 1.

We use AdamW with weight decay of $0.1$ and coefficients $\beta_1=0.9$ and $\beta_2=0.98$. The learning rate warms up to $1.0\times10^{-6}$ over the first 10\% of training, then decays to $1.0\times10^{-7}$ following a cosine schedule.

\section{Additional Dataset Analysis}
\label{app:dataset-analysis}

\subsection{Python Ecosystem Coverage}
\label{app:python-ecosystems}

\begin{figure}[h]
    \centering
    \includegraphics[width=\linewidth]{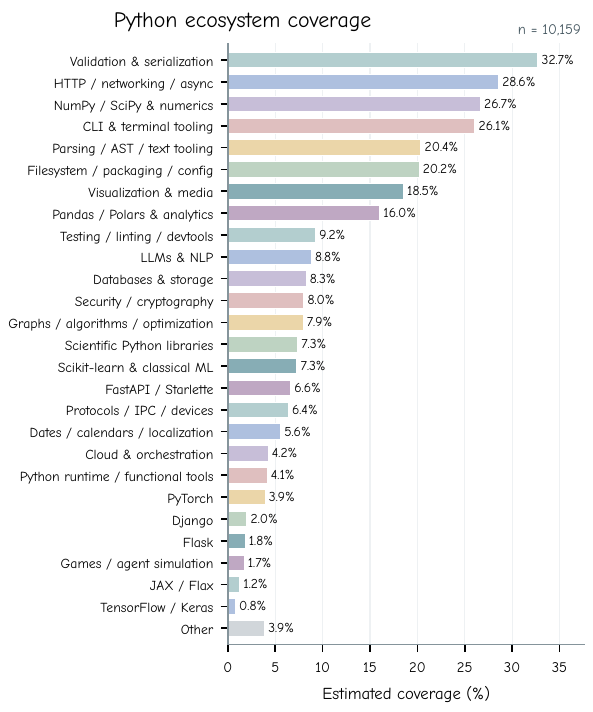}
    \caption{Estimated Python ecosystem coverage across RepoFactory's 10,159 environments. Categories may overlap. Other denotes environments without an ecosystem match.}
    \label{fig:python-ecosystems}
\end{figure}

Figure~\ref{fig:python-ecosystems} summarizes the estimated coverage of Python ecosystems across RepoFactory's 10,159 environments. The estimates use the available Python repository records. We assign ecosystem labels by matching direct runtime dependency names against predefined rules. For repositories without a dependency-based match, we use ecosystem terms in the first substantive paragraph of the project description. An environment may belong to multiple categories.

Validation and serialization are the most common category at 32.7\%, followed by HTTP, networking, and asynchronous tooling at 28.6\%. Numerical computing, command-line tools, text processing, and data analysis are also widely represented. The distribution extends from common application infrastructure to specialized scientific libraries, web frameworks, and machine-learning frameworks, covering a range of software stacks for repository reconstruction.

\subsection{Development Operations in Training Trajectories}
\label{app:development-operations}

\begin{figure}[h]
    \centering
    \includegraphics[width=\linewidth]{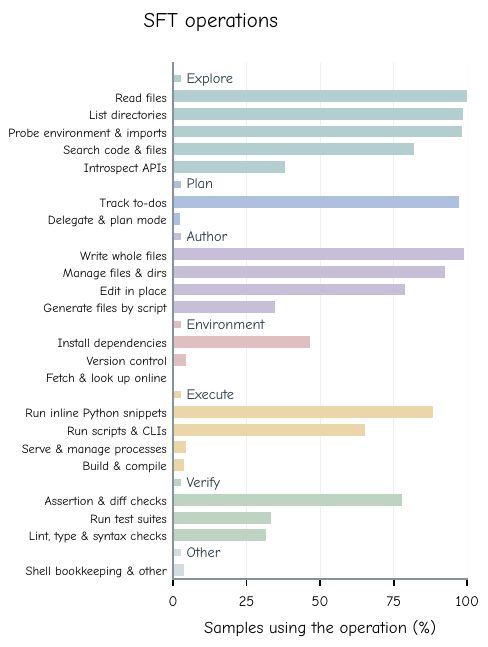}
    \caption{Development operations in the collected trajectories. Each bar shows the percentage of trajectories containing the operation. Categories may overlap.}
    \label{fig:tool-richness}
\end{figure}

We examine the development activities represented in the collected trajectories by classifying operations from tool names and shell-command contents. Figure~\ref{fig:tool-richness} reports the percentage of trajectories containing each operation. A trajectory may contribute to several categories.

Reading and writing files occur in almost all trajectories, alongside frequent workspace exploration and code modification. Inline Python execution appears in 88.2\% of trajectories, assertion or diff checks in 78.0\%, and test-suite execution in 33.2\%. These activities provide supervision for both implementing project code and examining its behavior through execution and local checks.

\end{document}